\documentclass[journal]{IEEEtran}
\usepackage{cite}
\usepackage{amsmath,amssymb,amsfonts}
\usepackage{bm}
\usepackage{graphicx}
\usepackage{algorithm,algorithmic}
\usepackage{booktabs}
\usepackage{multirow}
\usepackage{array}
\usepackage{hyperref}
\hypersetup{hidelinks=true}
\usepackage{pifont}
\newcommand{\cmark}{\ding{51}}
\newcommand{\xmark}{\ding{55}}
\usepackage{textcomp}
\def\BibTeX{{\rm B\kern-.05em{\sc i\kern-.025em b}\kern-.08em
    T\kern-.1667em\lower.7ex\hbox{E}\kern-.125emX}}

\begin{document}

\title{CardioDrift: Automatic Cardiac Motion Tracking on a 3D Atlas Mesh via Shape-aware Coherent Point Drift with Slice Shift Correction}

\author{Reza Akbari Movahed, Abuzar Rezaee, Shatha Alamri, Edmond S.L. Ho, Fani Deligianni, Ali Gooya
\thanks{This work has been submitted to the IEEE for possible publication. Copyright may be transferred without notice, after which this version may no longer be accessible.}
\thanks{This study used the UK Biobank Resource under application number 71392. The authors sincerely thank Dr Rona Strawbridge, School of Health and Wellbeing, University of Glasgow, for granting access to the dataset and for her valuable guidance on data usage and interpretation.}
\thanks{Reza Akbari Movahed,  Abuzar Rezaee, Shatha Alamri, Edmond S.L. Ho, Fani Deligianni, and Ali Gooya are with the School of Computing Science, University of Glasgow, Glasgow, United Kingdom (e-mail: 2919485a@student.gla.ac.uk; 3128738R@student.gla.ac.uk, 3071412A@student.gla.ac.uk, Shu-Lim.Ho@glasgow.ac.uk; Fani.Deligianni@glasgow.ac.uk; Ali.Gooya@glasgow.ac.uk)}}

\maketitle

\begin{abstract}
Automatic cardiac motion tracking from cine cardiac magnetic resonance (CMR) imaging is essential for quantifying myocardial motion abnormalities associated with cardiovascular disease. Existing methods fall into two groups: those that represent motion without anatomical correspondence across subjects and therefore limit inter-subject motion assessment; and those that represent motion with globally corresponding points, which typically require a time-consuming and error-prone semi-automatic step to fit a cardiac atlas to the subject space. To address these limitations, this study proposes CardioDrift, a fully automatic, atlas-based probabilistic framework that represents subject-specific cardiac motion on a 3D atlas mesh with globally corresponding points. The Coherent Point Drift (CPD) formulation is generalised to jointly register the atlas surfaces of multiple cardiac regions to the anatomically corresponding contours extracted from cine CMR segmentation maps throughout the cardiac cycle. Slice-shift correction is embedded in the same probabilistic model, estimating slice-wise translations that suppress the misalignment artefact limiting atlas-fitting accuracy. Evaluation on 1{,}301 randomly selected UK Biobank subjects and the M\&M dataset, CardioDrift attains Dice similarity coefficients (DSCs) of $92.23\pm8.70\%$, $82.11\pm7.56\%$, and $88.69\pm4.13\%$ for the left ventricle (LV), myocardium (Myo), and right ventricle (RV), exceeding the strongest baseline on every distance and overlap metric of both datasets.
\end{abstract}


\section{Introduction} \label{sec_1}
\IEEEPARstart{C}{ardiac} motion assessment plays a central role in the early detection of cardiovascular diseases, a leading cause of mortality worldwide \cite{claus2015tissue}. It is typically performed through visual assessment of cine cardiac magnetic resonance (CMR) imaging \cite{salerno2017recent}, a process that is subjective, poorly reproducible, and time-consuming \cite{luijnenburg2010intra}, motivating considerable effort to automate tracking and quantify motion abnormalities.

Automated methods face two obstacles. Those operating in the image domain represent motion in a subject-specific coordinate space, establishing no anatomical correspondence across individuals and precluding the population-level comparison that assessing pathological motion depends on. Those using a cardiac atlas mesh achieve such correspondence but require a semi-automatic step to fit the atlas to the subject's space, a process that is error-prone and time-consuming. Moreover, these methods first extract cardiac motion in image space, where it may include motion fields unrelated to the heart, and only then interpolate the motion fields from the image domain onto the mesh, attenuating them in the process.

This work addresses these obstacles with CardioDrift, a non-learning-based point set registration framework formulated using the Coherent Point Drift (CPD) approach \cite{myronenko2010point} that automatically fits a cardiac atlas mesh into the subject's cardiac space and propagates it over the cardiac cycle, using only contour points from cine short-axis (SAX) segmentation maps. A rigid stage first aligns the atlas at the end-diastolic (ED) frame while estimating per-slice translations to correct slice misalignment. A non-rigid stage then deforms the aligned atlas to the slice-shift-corrected contours at each subsequent frame through a shared deformation field. While previous methods lack global correspondence in the derived motion, rely on semi-automatic atlas fitting, capture irrelevant motion through image-domain motion extraction, and attenuate motion through interpolation, CardioDrift automatically fits a cardiac atlas to the subject space accurately and propagates it precisely throughout the cardiac cycle. The main contributions are:

\begin{itemize}
  \item Developing a probabilistic point-set registration framework for tracking cardiac motion on a 3D atlas mesh that provides anatomical correspondence across subjects, requiring only cine SAX segmentation maps as input.

  \item Generalising CPD to jointly register the left-ventricular endocardium (Endo), left-ventricular epicardium (Epi), and right-ventricular (RV) atlas surfaces to their corresponding contours using a single shared transformation.

  \item Embedding slice-shift correction in the rigid stage by estimating SAX-slice translations within the same probabilistic model to suppress inter-slice misalignment and improve atlas-fitting performance. 

  \item Evaluating the framework on the UK Biobank and Multi-Centre, Multi-Vendor and Multi-Disease (M\&M) datasets, demonstrating more accurate atlas fitting and propagation than existing methods, without manual alignment to the subject space.
\end{itemize}

\section{Background} \label{sec_bg}
Most state-of-the-art methods formulate tracking as image registration, estimating dense pixel- or voxel-wise motion fields from cine CMR. Classical approaches build on free-form deformation \cite{rueckert1999nonrigid, decraene2012temporal} or the Demons algorithm and its diffeomorphic reformulation \cite{thirion1998image, vercauteren2007non}, while deep-learning methods regress the field in a single forward pass \cite{balakrishnan2019voxelmorph, qin2018joint, meng2022mulvimotion}. Bayesian formulations register long-axis sequences \cite{zakeri2023dragnet} or jointly segment and register SAX slices \cite{bi2024segmorph}. They may capture motions unrelated to the heart because they rely on intensity-based image registration metrics. CardioMorphNet addresses this by guiding motion estimation towards the regions identified by a segmentation branch \cite{MOVAHED2026104149}. These methods operate in a subject-specific image space and establish no anatomical correspondence across individuals \cite{puyol2017multimodal, american2002standardized}, limiting inter-subject motion assessment.

Recent studies have formulated cardiac motion tracking in the mesh/point-set domain to enable both intra-subject and inter-subject motion assessment. For instance, Zakeri et al.\ proposed a deep motion model that predicts the cardiac point set at each time frame and identifies outlier points resulting from excessive deviations from the model prediction \cite{zakeri2022probabilistic}. Although this approach enables intra-subject assessment through outlier detection, its output point sets lack inter-subject correspondence, limiting abnormality assessment to the intra-subject setting. To enable inter-subject correspondence, DeepMesh fits an atlas mesh whose points correspond anatomically across individuals to the subject's cardiac space and warps it across the cycle using motion fields estimated from multi-view CMR \cite{meng2023deepmesh}. This method represents cardiac motion with globally corresponding points under several constraints: the fitting is semi-automatic and requires manual landmark selection; the motion fields are estimated in image space and interpolated onto the mesh in physical coordinates, attenuating extracted motion; and the intensity-based similarity term admits motions unrelated to the heart. Automating atlas fitting is complicated by the heart's variable position and orientation in physical space \cite{duan2019automatic} and by slice misalignment \cite{tarroni2020large}.

Point-set registration could enable atlas-based motion tracking without manual atlas fitting or image-to-mesh interpolation by operating directly in physical coordinates. Here, contours extracted from cine CMR segmentation maps serve as the fixed point sets, while the corresponding atlas surfaces are the moving point sets. Estimating transformations over the cardiac cycle that fit the atlas surfaces with the subject’s contours represents cardiac motion on the atlas mesh. Non-learning-based methods derive transformations by estimating point-wise correspondences through iterative optimisation or probabilistic modelling \cite{huang2021comprehensive}. Among these, CPD performs strongly \cite{myronenko2010point}, formulating registration as probability density estimation, where moving points parameterise a Gaussian mixture model (GMM) centroids and correspondences are expressed probabilistically and jointly refined with the transformation. Learning-based methods instead predict transformations from latent point-wise features, but generalise poorly beyond their training distribution and degrade under outliers, noise, and sparsity \cite{bojanic2024addressing}. These limitations are particularly relevant to CMR-derived contours, which are sparse through-plane, affected by slice misalignment, and prone to segmentation errors. CPD therefore offers a reliable alternative, but its standard formulation neither enforces correspondence between each atlas surface and its anatomical counterpart nor corrects slice misalignment, which limits fitting accuracy. 

\section{Materials and Methods}\label{sec_2}
\subsection{Proposed Framework}
The framework registers the Endo, Epi, and RV atlas surfaces to the subject's corresponding contour points throughout the cardiac cycle. Let \(g \in \{\text{Endo}, \text{Epi}, \text{RV}\}\) denote the cardiac anatomical region and \(l = 1, \dots, L\) the frames in the cardiac sequence, with \(l = 1\) denoting the ED time point. For each region, $\mathbf{X}^{g}_{l} \in \mathbb{R}^{N_{g} \times D}$ is the subject's contour point set at frame $l$ and $\mathbf{Y}^{g} \in \mathbb{R}^{M_{g} \times D}$ is the atlas point set, where $N_{g}$ and $M_{g}$ are their numbers of points and $D = 3$. Row-wise concatenation over the three regions in a fixed order, $\operatorname{Concat}_{g}(\cdot)$, gives $\mathbf{X}_{l} = \operatorname{Concat}_{g}(\mathbf{X}^{g}_{l}) \in \mathbb{R}^{N \times D}$ and $\mathbf{Y} = \operatorname{Concat}_{g}(\mathbf{Y}^{g}) \in \mathbb{R}^{M \times D}$, with $N = \sum_{g} N_{g}$ and $M = \sum_{g} M_{g}$. Given $\{\mathbf{X}_{l}\}_{l=1}^{L}$ and $\mathbf{Y}$, we aim to obtain $\{\mathbf{Z}_{l}\}_{l=1}^{L}$, where $\mathbf{Z}_{l} \in \mathbb{R}^{M \times D}$ is the registered atlas in the subject's cardiac space at frame $l$.

\begin{figure}[!t]
    \centering
    \includegraphics[width=0.5\textwidth]{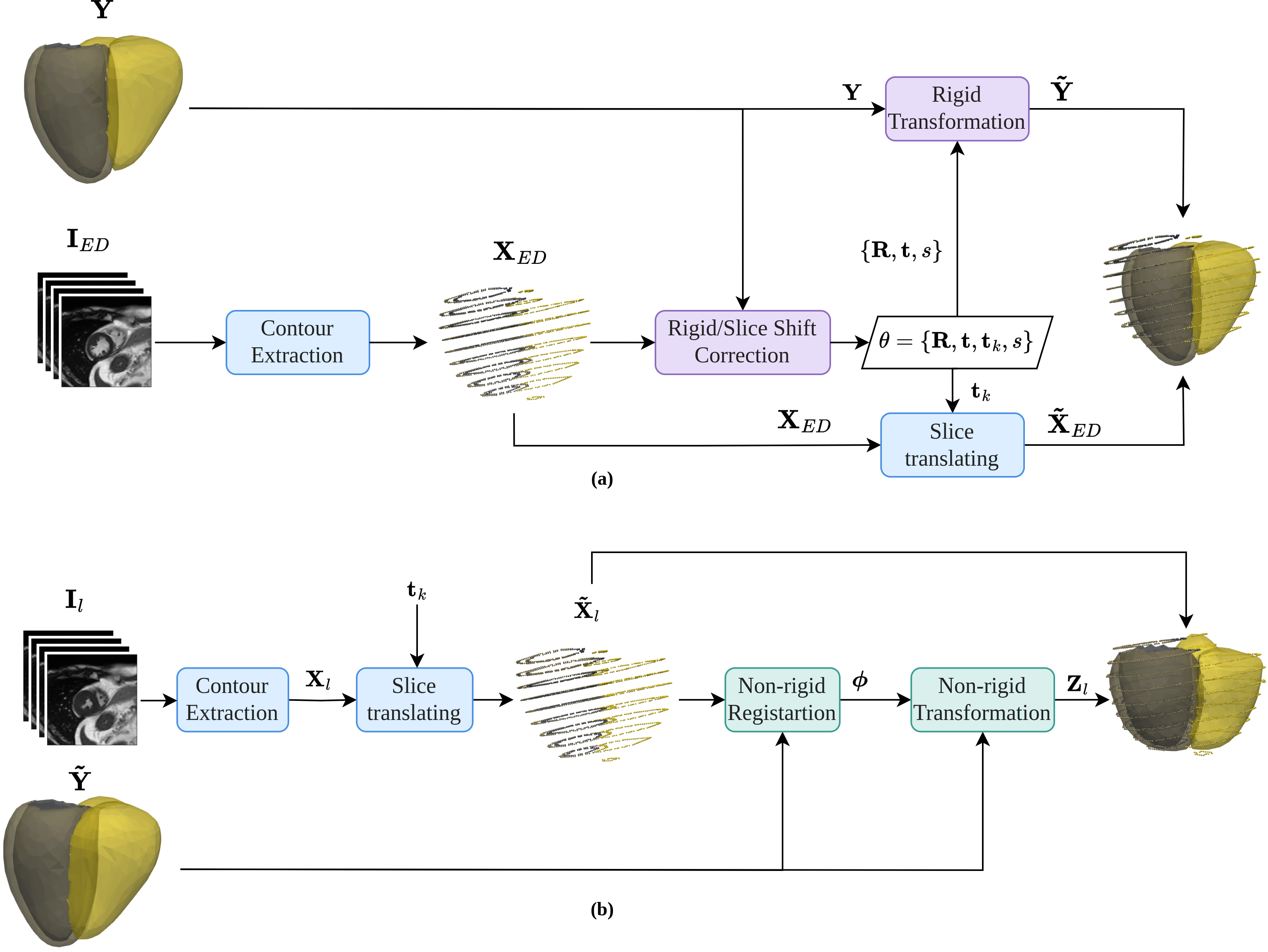}
    \caption{\fontsize{8pt}{10pt}\selectfont Overview of CardioDrift. (a) Alignment phase: the atlas $\mathbf{Y}$ is fitted to the subject's cardiac space at ED, which also estimates the per-slice translations. (b) Propagation phase: $\widetilde{\mathbf{Y}}$ is deformed to the corrected contours $\widetilde{\mathbf{X}}_l$ at each frame, providing $\mathbf{Z}_l$. $\{\mathbf{R}, \mathbf{t}, s\}$ are the rigid parameters, $\mathbf{t}_k$ are the slice shift correction translations, and $\boldsymbol{\phi}$ is the non-rigid deformation field.}
    \label{fig1}
\end{figure}

Fig.~\ref{fig1} shows the diagram of the proposed framework, whose input is a sequence of $L$ SAX volumes $\{\mathbf{I}_l\}_{l=1}^{L}$. Contour extraction proceeds identically at every frame: each SAX volume is segmented into the left ventricle (LV), myocardium (Myo), and RV, the Endo, Epi, and RV contours are extracted from the segmentation masks by the marching squares algorithm together with logical operations, and the result is mapped from voxel to physical coordinates using the affine transformation in the DICOM header, outputting $\mathbf{X}_{l}$.

Tracking proceeds in two phases. In the alignment phase (Fig.~\ref{fig1}a), the rigid parameters and the per-slice translations $\{\mathbf{t}_k \in \mathbb{R}^{D \times 1}\}_{k=1}^{K}$ over the $K$ SAX slices are estimated jointly from the atlas points $\mathbf{Y}$ and the ED contour points $\mathbf{X}_{\text{ED}}$; applying the rigid transformation to $\mathbf{Y}$ yields the fitted atlas $\widetilde{\mathbf{Y}}$ in the subject's cardiac space. In the propagation phase (Fig.~\ref{fig1}b), the contour points $\mathbf{X}_l$ at each frame are translated by the estimated slice-shift corrections to output $\widetilde{\mathbf{X}}_l$, and the non-rigid registration deforms $\widetilde{\mathbf{Y}}$ to $\widetilde{\mathbf{X}}_l$, yielding the frame-specific fitted atlas $\mathbf{Z}_l$. Repeating this over all $L$ frames provides $\{\mathbf{Z}_l\}_{l=1}^{L}$.

\subsection{Shape-aware CPD Generalisation}
Both stages build on the following generalisation of CPD, which registers the Endo, Epi, and RV atlas surfaces to their anatomically corresponding contour point sets under a single transformation. Following the original CPD formulation \cite{myronenko2010point}, registration is cast as probability density estimation: the atlas points of each region are the centroids of a GMM and the contour points are observations drawn from it, so that $p(\mathbf{x}_n^g) = \sum_{m=1}^{M_g+1}p(m^g)\,p(\mathbf{x}_n^g \mid m^g)$, where $\mathbf{x}^g_n \in \mathbb{R}^{1 \times D}$ is the $n$th contour point and $\mathbf{y}^g_m \in \mathbb{R}^{1 \times D}$ the $m$th atlas point of region $g$. Each component likelihood is an isotropic Gaussian,
\begin{equation}\label{eq2}
    p(\mathbf{x}_n^g \mid m^g) = \mathcal{N}\!\left(\mathbf{x}_n^g;\, \mathbf{y}_{m}^{g},\, \sigma^{2}_g\mathbf{I}\right),
\end{equation}
with equal priors $p(m^g) = 1/M_g$ and an additional uniform component $p(\mathbf{x}_n^g \mid M_g+1) = 1/N_g$, weighted by $\omega_g\in (0,1)$, accounting for outliers.

The GMM centroids are re-parameterised by a transformation parameter set $\theta$, estimated by minimising the weighted negative log-likelihood
\begin{equation}\label{eq4}
    E(\theta, \sigma^{2}_g) = -\sum_{g}\alpha_g\sum_{n=1}^{N_g} \log \sum_{m=1}^{M_g+1} p(m^g)\,p(\mathbf{x}_n^g\mid m^g),
\end{equation}
where $\alpha_g \in [0,1]$ weights region $g$. The posterior that the $m$th component generated the $n$th observation, $P^{g}_{mn} \triangleq p(m^g \mid \mathbf{x}_n^g)$ representing the soft correspondence between $\mathbf{x}_n^g$ and $\mathbf{y}_m^g$, follows from Bayes' rule as
\begin{equation}\label{eq5}
\small
P^{g}_{mn} =\frac{\exp\!\left(-\dfrac{1}{2\sigma^{2}_g}\lVert \mathbf{x}_n^g- \tau(\mathbf{y}_{m}^{g};\, \theta)\rVert^2\right)}{\displaystyle\sum_{j=1}^{M_g}\exp\!\left(-\dfrac{1}{2\sigma^{2}_g}\lVert \mathbf{x}_n^g- \tau(\mathbf{y}_{j}^{g};\, \theta)\rVert^2\right)+c},
\end{equation}
where $\tau(\,\cdot\,;\theta)$ is the transformation parameterised by $\theta$ and $c = (2\pi\sigma^{2}_g)^{D/2}\,\frac{w_g}{1-w_g}\,\frac{M_g}{N_g}$.

The optimal $\theta$ and $\sigma^{2}_g$ are derived by expectation-maximisation (EM): the E-step computes $P^{g}_{mn}$ from the current estimates, and the M-step minimises the expected complete-data negative log-likelihood evaluated at the previous values $P^{g(\text{old})}_{mn}$, which upper-bounds $E$. Dropping terms independent of $\theta$ and $\sigma^{2}_g$,
\begin{equation}\label{eq7}
\begin{split}
    Q(\theta,\sigma^{2}_g) = \sum_{g}\alpha_g\bigg( & \frac{1}{2\sigma^{2}_g}\sum_{n,m} P^{g(\text{old})}_{mn} \lVert \mathbf{x}_n^g - \tau(\mathbf{y}_{m}^{g};\theta)\rVert^2 \\
    & + \frac{N^g_p D}{2}\log\sigma^{2}_g \bigg),
\end{split}
\end{equation}
where $N^g_p = \sum_{n,m}P^{g(\text{old})}_{mn}$. The form of $\tau(\,\cdot\,;\theta)$ is given separately for the rigid and non-rigid stages below.

\subsection{Rigid and Slice-Shift-Correction Transformation}
This stage fits the atlas surfaces to their corresponding contours while estimating per-slice translations. The transformation is
\begin{equation}\label{eq_tau_rigid}
    \tau\!\left(\mathbf{y}^g_m;\, \mathbf{t}_k, \mathbf{R}, \mathbf{t}, s\right) = \mathbf{t}_k + s\mathbf{R}\mathbf{y}^g_m + \mathbf{t},
\end{equation}
where $\mathbf{x}_{nk}^g$ is the $n$th contour point of region $g$ on the $k$th SAX slice, $\mathbf{t}_k$ denotes slice-shift translations for the point on the $k$th slice. $\mathbf{R} \in \mathbb{R}^{D \times D}$, $\mathbf{t} \in \mathbb{R}^{D \times 1}$, and $s \in \mathbb{R}^{+}$ are the rotation, global translation, and isotropic scaling. Substituting into \eqref{eq7} gives
\begin{equation}\label{eq8}
\begin{split}
    Q(\mathbf{t}_k, \mathbf{R}, \mathbf{t}, s, \sigma^{2}_g) =  & \sum_{g}\alpha_g\bigg(\frac{1}{2\sigma^{2}_g}\sum_{k,n,m} P^{g(\text{old})}_{mnk} \\
    & \hspace{-8em} \times \lVert \mathbf{x}_{nk}^g - \mathbf{t}_k - s\mathbf{R}\mathbf{y}^g_m - \mathbf{t} \rVert^2 + \frac{N^g_p D}{2}\log\sigma^{2}_g\bigg),
\end{split}
\end{equation}
the sum runs over $k = 1,\dots,K$, $n = 1,\dots,N^k_g$, and $m = 1,\dots,M_g$ with $N^k_g$ the number of contour points of region $g$ on slice $k$, and $N^g_p = \sum_{k,n,m}P^{g(\text{old})}_{mnk}$. The corresponding posterior is
\begin{equation}\label{eq9}
\small
    P^{g}_{mnk}=\frac{\exp\!\left(-\dfrac{1}{2\sigma^{2}_g}\lVert \mathbf{x}_{nk}^g - \mathbf{t}_k - s\mathbf{R}\mathbf{y}^g_m - \mathbf{t}\rVert^2\right)}{\displaystyle\sum_{j=1}^{M_g}\exp\!\left(-\dfrac{1}{2\sigma^{2}_g}\lVert \mathbf{x}_{nk}^g - \mathbf{t}_k - s\mathbf{R}\mathbf{y}^g_j - \mathbf{t}\rVert^2\right)+c}.
\end{equation}

Minimising \eqref{eq8} jointly over $\{\mathbf{R}, \mathbf{t}, s, \sigma^{2}_g, \mathbf{t}_k\}$ is intractable, as the rigid parameters and slice-shift translations enter the objective in a coupled manner. We therefore alternate two stages within each EM iteration: the rigid parameters are updated on contour points translated by the current $\mathbf{t}_k$, then $\mathbf{t}_k$ is re-estimated from the residual misalignment between the warped atlas $\widetilde{\mathbf{Y}}^g$ and the original contours $\mathbf{X}^g$. Both derivations follow below.

\subsection{Rigid Parameters Estimation}
This step estimates $\mathbf{R}$, $\mathbf{t}$, and $s$ from the atlas point sets and the contour points after the current slice-shift translation has been applied. Writing $\tilde{\mathbf{x}}^g_{n} = \mathbf{x}^g_{nk} - \mathbf{t}_k$, the objective \eqref{eq8} reduces to \eqref{eq7} under $\tau(\mathbf{y}^g_m) = s\mathbf{R}\mathbf{y}^g_m + \mathbf{t}$ with $\widetilde{\mathbf{X}}^g$ in place of $\mathbf{X}^g$, the posteriors $P^{g}_{mn}$ following from \eqref{eq5} under the same substitutions. Minimising this objective over $\mathbf{R}$ is constrained by $\mathbf{R}^\top\mathbf{R} = \mathbf{I}$ and $\det(\mathbf{R}) = 1$, and admits a closed form via Lemma~1 of \cite{myronenko2009closed}: the rotation maximising $\operatorname{tr}(\mathbf{A}^\top\mathbf{R})$ for a known real $\mathbf{A} \in \mathbb{R}^{D \times D}$ is $\mathbf{R} = \mathbf{U}\mathbf{C}\mathbf{V}^\top$, with $\mathbf{U}\mathbf{S}\mathbf{V}^\top = \operatorname{SVD}(\mathbf{A})$, $\mathbf{C} = \operatorname{diag}(1, \ldots, \det(\mathbf{U}\mathbf{V}^\top))$, and $\operatorname{diag}(\cdot)$, $\operatorname{tr}(\cdot)$ the diagonal matrix formed from its arguments and the trace.

To reach this form, $\mathbf{t}$ is eliminated by setting $\partial Q / \partial
\mathbf{t} = 0$, giving
\begin{equation}\label{eq12}
    \mathbf{t} = \sum_{g} w_g\!\left(\boldsymbol{\mu}^g_x - s\mathbf{R}\boldsymbol{\mu}^g_y\right)
    = \bar{\boldsymbol{\mu}}_x - s\mathbf{R}\bar{\boldsymbol{\mu}}_y,
\end{equation}
where $\boldsymbol{\mu}^g_x = \frac{1}{N^g_p}(\widetilde{\mathbf{X}}^g)^\top(\mathbf{P}^g)^\top\mathbf{1}$ and  $\boldsymbol{\mu}^g_y = \frac{1}{N^g_p}(\mathbf{Y}^g)^\top\mathbf{P}^g\mathbf{1}$ $\mathbf{P}^g \in \mathbb{R}^{M_g \times N_g}$ collects the posteriors $P^g_{mn}$, and $\mathbf{1}$ is a column vector of ones of
the appropriate length. The region weights
\begin{equation}\label{eq_wg}
    w_g = \frac{\gamma_g N^g_p}{\sum_{h}\gamma_h N^h_p},
    \qquad \gamma_g \triangleq \frac{\alpha_g}{\sigma^{2}_g},
\end{equation}
sum to one over the three regions, $h$ indexing the same set as $g$, so that
$\mathbf{t}$ is the offset between the pooled centroids
$\bar{\boldsymbol{\mu}}_x = \sum_{g} w_g \boldsymbol{\mu}^g_x$ and
$\bar{\boldsymbol{\mu}}_y = \sum_{g} w_g \boldsymbol{\mu}^g_y$.

Writing $\overline{\mathbf{X}}^g = \widetilde{\mathbf{X}}^g - \mathbf{1}\bar{\boldsymbol{\mu}}_x^\top$ and $\overline{\mathbf{Y}}^g = \mathbf{Y}^g - \mathbf{1}\bar{\boldsymbol{\mu}}_y^\top$, and substituting \eqref{eq12} back leaves no region-wise residual offset and reduces $Q$ to $c_1 \operatorname{tr}(\mathbf{A}^\top\mathbf{R}) + c_2$, with $c_1 > 0$, $c_2$ independent of $\mathbf{R}$, and
\begin{equation}\label{eq_A}
    \mathbf{A} = \sum_{g}\gamma_g \left(\overline{\mathbf{X}}^g\right)^\top
    \left(\mathbf{P}^g\right)^\top \overline{\mathbf{Y}}^g.
\end{equation}
Minimising $Q$ over $\mathbf{R}$ is therefore equivalent to maximising
$\operatorname{tr}(\mathbf{A}^\top\mathbf{R})$, and the optimal rotation follows
from Lemma~1. Setting the remaining partial derivatives to zero gives
\begin{equation}\label{eq_s_sigma}
\small
\begin{split}
    s &= \frac{\operatorname{tr}(\mathbf{A}^\top\mathbf{R})}
    {\sum_g \gamma_g\operatorname{tr}\!\left[(\overline{\mathbf{Y}}^g)^\top
    \operatorname{diag}(\mathbf{P}^g\mathbf{1})\overline{\mathbf{Y}}^g\right]}, \\
    \sigma^2_g &= \frac{1}{D N^g_p}\Big(
    \operatorname{tr}\!\left[(\overline{\mathbf{X}}^g)^\top
    \operatorname{diag}((\mathbf{P}^g)^\top\mathbf{1})\overline{\mathbf{X}}^g\right]
    - 2s\operatorname{tr}(\mathbf{A}_g^\top\mathbf{R}) \\
    &\qquad\qquad + s^2\operatorname{tr}\!\left[(\overline{\mathbf{Y}}^g)^\top
    \operatorname{diag}(\mathbf{P}^g\mathbf{1})\overline{\mathbf{Y}}^g\right]\Big),
\end{split}
\end{equation}
where $\mathbf{A}_g =(\overline{\mathbf{X}}^g)^\top(\mathbf{P}^g)^\top \overline{\mathbf{Y}}^g$. Each EM iteration of this stage computes $\mathbf{P}^g$ via \eqref{eq5}, forms $\gamma_g$ from the current $\sigma^{2}_g$, then applies \eqref{eq12}--\eqref{eq_s_sigma} in closed form.

\subsection{Slice-wise Translation Estimation}
This step estimates the per-slice translations $\{\mathbf{t}_k\}_{k=1}^{K}$, given the rigidly warped atlas points $\widetilde{\mathbf{y}}^g_{m} = s\mathbf{R}\mathbf{y}^g_m + \mathbf{t}$ and the original contour points $\mathbf{X}^g$. Holding the rigid parameters fixed, \eqref{eq8} reduces to
\begin{equation}\label{eq15}
    Q(\mathbf{t}_k) = \sum_{g}\frac{\alpha_g}{2\sigma^{2}_g}\sum_{k,n,m} P^{g(\text{old})}_{mnk} \lVert \mathbf{x}_{nk}^g - \mathbf{t}_k - \widetilde{\mathbf{y}}^g_{m} \rVert^2,
\end{equation}
with $P^{g(\text{old})}_{mnk}$ computed via \eqref{eq9}.

Each $\mathbf{t}_k$ must lie in the SAX imaging plane. The plane orientation is estimated from the data: the contour points on the middle SAX slice $\{\mathbf{p}_r\}_{r=1}^{N_m} ~\mathbf{p}_r \in \mathbb{R}^{1 \times D}$ where $N_m$ is the total number of such points, are centred to give $\widetilde{\mathbf{p}}_l$. Then, the eigenvectors of the scatter matrix $\mathbf{H} = \sum_{r}\widetilde{\mathbf{p}}_r^\top\widetilde{\mathbf{p}}_r\in \mathbb{R}^{D \times D}$ are obtained by SVD. The eigenvector of smallest eigenvalue $\mathbf{v}_1$ approximates the slice-stack normal, while the remaining $\mathbf{v}_2, \ldots, \mathbf{v}_D$ span the imaging plane and form the projection matrix $\mathbf{E} = [\mathbf{v}_2, \ldots, \mathbf{v}_D]^\top \in \mathbb{R}^{(D-1) \times D}$. Each translation is then parameterised as $\mathbf{t}_k = \mathbf{E}^\top \mathbf{a}_k$ with $\mathbf{a}_k \in \mathbb{R}^{(D-1) \times 1}$, guaranteeing $\mathbf{t}_k$ lies in the estimated plane.


Substituting this parameterisation into \eqref{eq15}, retaining only the terms in $\mathbf{a}_k$, and adding the regularisation term $\frac{\lambda}{2}\sum_{k}\mathbf{a}_k^\top\mathbf{a}_k$ that prevents unrealistically large translations, where $\lambda > 0$ is the regularisation coefficient, the stationary condition $\partial Q / \partial \mathbf{a}_k = 0$ together with $\mathbf{t}_k = \mathbf{E}^\top\mathbf{a}_k$ yields 
\begin{equation}\label{eq19}
    \mathbf{t}_k = \frac{\mathbf{E}^\top\mathbf{E}\displaystyle\sum_{g,n,m}\frac{\alpha_g}{\sigma^2_g}P^{g(\text{old})}_{mnk}\left(\mathbf{x}_{nk}^g - \widetilde{ \mathbf{y}}^g_{m}\right)}{\displaystyle\sum_{g,n,m}\frac{\alpha_g}{\sigma^2_g}P^{g(\text{old})}_{mnk} + \lambda},
\end{equation}
applied in the M-step once $P^{g(\text{old})}_{mnk}$ has been computed in the E-step.

\subsection{Non-rigid Transformation}
The non-rigid stage deforms the fitted atlas to the contours at each frame. With the fitted atlas points $\widetilde{\mathbf{Y}} = \operatorname{Concat}_{g}(\widetilde{\mathbf{Y}}^g)$, the deformed point set is $\mathbf{Z} = \widetilde{\mathbf{Y}} + \boldsymbol{\phi}(\widetilde{\mathbf{Y}})$, where $\boldsymbol{\phi}$ is the deformation field applied uniformly across all regions. Following \cite{myronenko2010point}, $\boldsymbol{\phi}(\widetilde{\mathbf{Y}}) = \mathbf{G}\mathbf{W}$, with $\mathbf{W} \in \mathbb{R}^{M \times D}$ an unknown weight matrix and $\mathbf{G} \in \mathbb{R}^{M \times M}$ a Gaussian kernel matrix constructed by,
\begin{equation}\label{eq_kernel}
G_{ij} = \exp\left(-\frac{\lVert\mathbf{y}_i - \mathbf{y}_j\rVert^2}{2\rho^2}\right), \quad i, j= 1, \dots, M, 
\end{equation}
whose bandwidth $\rho > 0$ controls the spatial extent of the deformation. Substituting $\tau(\widetilde{\mathbf{Y}}, \mathbf{W}) = \widetilde{\mathbf{Y}} + \mathbf{G}\mathbf{W}$ into \eqref{eq7} and adding the penalty $\frac{\beta}{2}\operatorname{tr}[\mathbf{W}^\top\mathbf{G}\mathbf{W}]$, in which $\beta > 0$ governs the smoothness of the deformation field, specialises the objective to a function $Q(\mathbf{W}, \sigma^{2}_g)$, with the posteriors $\mathbf{P}^g$ following from \eqref{eq5} under this $\tau$. Writing $\mathbf{G}(m,\,\colon)$ for the $m$th row of $\mathbf{G}$, setting $\partial Q / \partial \mathbf{W} = 0$ and $\partial Q / \partial \sigma^2_g = 0$ yields, as shown in \cite{myronenko2010point},
\begin{equation}\label{eq21}
    \left(\mathbf{G} + \beta\,\mathbf{D}_\sigma\right)\mathbf{W} = \operatorname{diag}(\mathbf{P}\mathbf{1})^{-1}\mathbf{P}\mathbf{X} - \widetilde{\mathbf{Y}},
\end{equation}
\begin{equation}\label{eq23}
    \sigma^2_g = \frac{1}{N^g_p D}\sum_{n,m} P^{g(\text{old})}_{mn}\left\lVert\tilde{\mathbf{x}}_n^g - \tilde{\mathbf{y}}^g_m - \mathbf{G}(m,\,\colon)\mathbf{W}\right\rVert^2,
\end{equation}
where $\widetilde{\mathbf{X}} = \operatorname{Concat}_{g}(\widetilde{\mathbf{X}}^g)$, $\mathbf{P} = \operatorname{blkdiag}_{g}(\mathbf{P}^g)$, and $\mathbf{D}_{\sigma} = \operatorname{blkdiag}_g(\sigma^{2}_{g}/\alpha_g \operatorname{diag}(\mathbf{P}^{g}\mathbf{1})^{-1})$. Initialised with $\mathbf{W} = \mathbf{0}$ and $\sigma_g^2 = \frac{1}{D N_g M_g}\sum_{n,m}\lVert \tilde{\mathbf{x}}^g_n - \mathbf{y}^g_m \rVert^2$, the EM alternates between computing $P^{g(\text{old})}_{mn}$ via \eqref{eq5} and solving \eqref{eq21}--\eqref{eq23} until convergence.

\section{Experiments and Results}
\subsection{Experimental Setup}
The framework was evaluated on two datasets. The first comprises cine CMR SAX volumes from a random sample of 1{,}301 UK Biobank subjects \cite{petersen2015uk}, segmented into the LV, Myo, and RV regions at every frame of the cardiac cycle with the automated tool of Bai et al.\ \cite{bai2018automated}. The second is the M\&M dataset \cite{campello2021multi}, which includes 134 subjects; it was acquired at several clinical centres with scanners from four vendors and includes both healthy and pathological anatomy. Expert-annotated masks are provided for M\&M only at the ED and end-systolic (ES) time points. The bi-ventricular atlas of \cite{bai2015bi} served as the template, with the Endo, Epi, and RV surfaces decimated to 359, 459, and 402 points, respectively.

The rigid stage is compared against two classical methods, iterative closest point (ICP) and rigid CPD, and two learning-based methods, PCRNet \cite{sarode2019pcrnet} and PointNetLK \cite{aoki2019pointnetlk}. The non-rigid stage is compared against non-rigid CPD, CPD-Net \cite{wang2019coherent}, a learning-based extension of CPD, and DeepMesh \cite{meng2023deepmesh}, which estimates motion fields from the cine images and interpolates them onto the mesh vertices. All non-rigid methods deform the atlas fitted by CardioDrift's rigid stage and receive contour points pre-translated by its slice-shift corrections, so the comparison isolates the deformation model; for DeepMesh, this replaces the semi-automatic initialisation of the original formulation. Ablations of the slice-shift correction and the shape-aware modelling are also reported. Both regularisation coefficients were selected by sweeps on the two datasets, given in full in the supplementary material. We adopt $\lambda = 5\times10^3$, which maximises UK Biobank accuracy at a mean per-slice shift of $2.6$~mm, within the range reported by Tarroni et al.\ \cite{tarroni2020large}, and $\beta = 3.16$, which balances accuracy against the mean Laplacian of the registered mesh; both are used unchanged for the two datasets.  

Performance is quantified by the mean surface distance (MSD) and the 95th-percentile Hausdorff distance (HD95), both in millimetres, computed between the fitted atlas surfaces and the subject contour points, and by the Dice similarity coefficient (DSC) between the reference masks and those obtained by rasterising the fitted surfaces onto the SAX volume grid. The rigid stage is evaluated at the ED frame, at which the atlas is fitted; the non-rigid stage is evaluated across all frames on UK Biobank and, following annotation availability, at ED and ES only on M\&M. 

\subsection{Rigid Registration Performance}
The UK Biobank rigid block of Table~\ref{tab:results} reports the quantitative comparison on that cohort. Against the baselines, CardioDrift achieves the lowest HD95 and MSD for all three regions and the highest DSC for the LV, Myo, and RV. CPD outperforms ICP on every metric but still yields substantially larger distances and lower overlap than CardioDrift, while ICP obtains the largest geometric errors. PCRNet improves on PointNetLK in HD95 and MSD and attains a competitive RV DSC of 75.04\%, yet both lag behind throughout, reflecting the difficulty of deriving correspondences in a single forward pass from sparse and heterogeneous contours.

The same comparison on M\&M is reported in Table~\ref{tab:results}. The ordering of the methods is preserved, with CPD again the strongest baseline and ICP and PointNetLK the weakest, but absolute performance is lower for every method, while the margin of CardioDrift over the strongest baseline holds up, at $28.2$ percentage points of LV DSC and $28.3$ points of Myo DSC against $22.6$ and $28.5$ on UK Biobank, and remains substantial on all nine metrics. The learning-based methods degrade most in relative terms: PointNetLK loses $15.3$ percentage points of RV DSC ($51.48\%$ to $36.16\%$) and PCRNet $11.0$ points ($75.04\%$ to $64.06\%$). This is consistent with the domain-shift sensitivity of single-pass learned registration.

Fig.~\ref{fig_re}(a) shows a representative UK Biobank case. The surfaces fitted by CardioDrift closely follow the contour points across all three regions, and the rasterised masks closely match the corresponding regions in the ground-truth masks, whereas the competing methods show noticeable misalignment at the ventricular boundaries and in overall orientation.

\begin{table*}[!t]
    \centering
    \caption{CardioDrift against baseline registration methods on both datasets, with the slice-shift-correction (SSC) ablation. HD95 and MSD (mm) are computed against the subject contour points, DSC (\%) against the reference masks. On UK Biobank, the rigid stage is evaluated at ED and the non-rigid stage across all frames; on M\&M, the rigid stage is evaluated at ED and the non-rigid stage at ED and ES. \textbf{Bold}: best per column within each stage of each dataset.}
    \renewcommand{\arraystretch}{1.3}
    \resizebox{\textwidth}{!}{%
    \begin{tabular}{c|c l c c c c c c c c c}
        \hline\hline
        \multirow{2}{*}{Dataset} & \multirow{2}{*}{Stage} & \multirow{2}{*}{Method} & \multicolumn{3}{c}{HD95 (mm)} & \multicolumn{3}{c}{MSD (mm)} & \multicolumn{3}{c}{DSC (\%)}\\
        \cline{4-12}
        {} & {} & {} & Endo & Epi & RV & Endo & Epi & RV & LV & Myo & RV \\
        \hline
        \multirow{10}{*}{UK Biobank} & \multirow{6}{*}{Rigid}
        & CardioDrift &
            \textbf{14.18}$\pm$4.43 & \textbf{16.98}$\pm$5.06 & 22.30$\pm$4.78 &
            \textbf{4.50}$\pm$1.21 & \textbf{4.77}$\pm$1.39 & 5.72$\pm$1.52 &
            \textbf{90.72}$\pm$10.56 & \textbf{70.25}$\pm$12.27 & 81.01$\pm$9.28 \\
        & & CardioDrift w/o SSC &
            17.19$\pm$6.94 & 18.93$\pm$7.42 & \textbf{21.16}$\pm$4.83 &
            5.62$\pm$2.60 & 5.80$\pm$2.48 & \textbf{4.96}$\pm$1.16 &
            86.52$\pm$11.90 & 55.76$\pm$12.68 & \textbf{86.19}$\pm$8.82 \\
        & & CPD &
            21.45$\pm$11.30 & 23.09$\pm$11.12 & 32.66$\pm$15.98 &
            7.51$\pm$3.88 & 8.11$\pm$4.71 & 10.29$\pm$6.58 &
            68.13$\pm$29.96 & 41.76$\pm$27.89 & 57.96$\pm$31.88 \\
        & & ICP &
            38.10$\pm$6.46 & 36.75$\pm$6.28 & 44.35$\pm$6.42 &
            11.00$\pm$1.64 & 10.76$\pm$1.74 & 15.70$\pm$2.09 &
            34.19$\pm$17.39 & 11.70$\pm$5.01 & 50.32$\pm$5.94 \\
        & & PCRNet &
            24.79$\pm$4.62 & 25.37$\pm$5.08 & 40.29$\pm$6.85 &
            8.50$\pm$1.56 & 7.71$\pm$1.82 & 9.93$\pm$1.54 &
            56.01$\pm$14.77 & 25.03$\pm$11.18 & 75.04$\pm$5.76 \\
        & & PointNetLK &
            36.38$\pm$8.67 & 34.47$\pm$8.60 & 42.85$\pm$7.59 &
            9.28$\pm$1.74 & 8.64$\pm$1.87 & 14.61$\pm$2.66 &
            20.87$\pm$19.01 & 13.66$\pm$8.62 & 51.48$\pm$10.47 \\
        \cline{2-12}
        & \multirow{4}{*}{Non-rigid}
        & CardioDrift &
            \textbf{9.47}$\pm$3.14 & \textbf{10.62}$\pm$3.26 & \textbf{14.33}$\pm$2.92 &
            \textbf{3.01}$\pm$0.35 & \textbf{3.06}$\pm$0.35 & \textbf{2.89}$\pm$0.33 &
            \textbf{92.23}$\pm$8.70 & \textbf{82.11}$\pm$7.56 & \textbf{88.69}$\pm$4.13 \\
        & & CPD &
            13.73$\pm$4.87 & 12.12$\pm$5.26 & 17.31$\pm$5.28 &
            4.10$\pm$1.18 & 4.15$\pm$1.30 & 4.22$\pm$2.29 &
            85.23$\pm$13.02 & 64.91$\pm$9.98 & 83.15$\pm$10.51 \\
        & & CPD-Net &
            12.94$\pm$4.91 & 14.31$\pm$5.49 & 19.45$\pm$4.90 &
            4.58$\pm$1.63 & 4.69$\pm$1.71 & 5.16$\pm$2.08 &
            84.40$\pm$12.41 & 60.41$\pm$12.57 & 79.71$\pm$12.35 \\
        & & DeepMesh &
            17.56$\pm$5.46 & 19.88$\pm$6.02 & 19.92$\pm$4.31 &
            5.91$\pm$1.69 & 5.84$\pm$1.57 & 5.27$\pm$1.20 &
            81.19$\pm$12.25 & 60.90$\pm$15.79 & 76.46$\pm$10.96 \\
        \hline
        \multirow{10}{*}{M\&M} & \multirow{6}{*}{Rigid}
        & CardioDrift &
            \textbf{17.23}$\pm$6.98 & \textbf{22.15}$\pm$7.29 & 27.00$\pm$8.62 &
            \textbf{5.50}$\pm$2.36 & \textbf{6.19}$\pm$2.72 & 7.35$\pm$3.66 &
            \textbf{83.82}$\pm$20.34 & \textbf{60.21}$\pm$19.10 & \textbf{75.01}$\pm$18.54 \\
        & & CardioDrift w/o SSC &
            21.96$\pm$9.30 & 25.61$\pm$9.50 & \textbf{25.10}$\pm$7.85 &
            8.28$\pm$4.24 & 8.16$\pm$4.06 & \textbf{7.16}$\pm$2.31 &
            70.86$\pm$22.19 & 44.45$\pm$18.56 & 67.03$\pm$19.70 \\
        & & CPD &
            24.76$\pm$10.62 & 27.85$\pm$9.76 & 37.64$\pm$18.30 &
            8.53$\pm$3.73 & 9.51$\pm$4.51 & 12.20$\pm$6.80 &
            55.67$\pm$33.68 & 31.88$\pm$26.19 & 48.01$\pm$30.61 \\
        & & ICP &
            37.38$\pm$8.44 & 35.45$\pm$7.97 & 45.07$\pm$8.73 &
            10.80$\pm$2.26 & 10.86$\pm$2.21 & 15.66$\pm$2.80 &
            26.11$\pm$18.24 & 9.03$\pm$7.02 & 46.92$\pm$7.59 \\
        & & PCRNet &
            27.44$\pm$5.79 & 28.95$\pm$6.37 & 36.52$\pm$8.59 &
            8.88$\pm$2.04 & 8.06$\pm$2.26 & 10.71$\pm$2.47 &
            44.92$\pm$21.08 & 22.69$\pm$11.19 & 64.06$\pm$10.54 \\
        & & PointNetLK &
            33.45$\pm$7.83 & 31.32$\pm$7.79 & 47.21$\pm$10.74 &
            12.58$\pm$2.30 & 11.09$\pm$2.13 & 16.94$\pm$4.39 &
            21.32$\pm$17.73 & 10.86$\pm$6.91 & 36.16$\pm$10.53 \\
        \cline{2-12}
        & \multirow{4}{*}{Non-rigid}
        & CardioDrift &
            \textbf{10.95}$\pm$4.32 & \textbf{13.40}$\pm$4.55 & \textbf{14.57}$\pm$3.55 &
            \textbf{3.12}$\pm$0.64 & \textbf{3.35}$\pm$0.69 & \textbf{2.66}$\pm$0.46 &
            \textbf{86.66}$\pm$14.22 & \textbf{76.81}$\pm$12.43 & \textbf{84.53}$\pm$9.33 \\
        & & CPD &
            15.59$\pm$8.21 & 16.08$\pm$7.65 & 20.46$\pm$9.75 &
            5.16$\pm$2.24 & 5.14$\pm$2.31 & 5.33$\pm$3.78 &
            73.13$\pm$24.15 & 55.80$\pm$16.18 & 74.94$\pm$21.52 \\
        & & CPD-Net &
            17.65$\pm$9.22 & 20.81$\pm$8.78 & 22.21$\pm$7.86 &
            6.26$\pm$3.05 & 6.98$\pm$3.24 & 6.65$\pm$3.78 &
            73.61$\pm$22.50 & 48.35$\pm$17.29 & 68.75$\pm$21.69 \\
        & & DeepMesh &
            28.76$\pm$11.04 & 31.79$\pm$10.58 & 30.48$\pm$10.81 &
            9.22$\pm$3.71 & 9.45$\pm$3.57 & 8.76$\pm$2.41 &
            66.04$\pm$22.25 & 38.04$\pm$14.87 & 57.88$\pm$18.23 \\
        \hline
    \end{tabular}
    }
    \label{tab:results}
\end{table*}

\subsection{Non-rigid Registration Performance}
The UK Biobank non-rigid block of Table~\ref{tab:results} reports atlas propagation over the cardiac cycle. CardioDrift achieves the lowest HD95 and MSD for all regions and the highest DSC throughout. The largest gain is in the Myo DSC, which rises from $64.91\pm9.98\%$ (CPD), $60.41\pm12.57\%$ (CPD-Net), and $60.90\pm15.79\%$ (DeepMesh) to $82.11\pm7.56\%$.  DeepMesh records the largest surface distances of the four methods, an Endo HD95 of $17.56\pm5.46$~mm against $9.47\pm3.14$~mm for CardioDrift.

The M\&M non-rigid block reports the corresponding results, which are confined to ED and ES. CardioDrift is best on every metric, and the Myo again shows the widest margin: $76.81\pm12.43\%$ against $55.80\pm16.18\%$ for CPD, $48.35\pm17.29\%$ for CPD-Net, and $38.04\pm14.87\%$ for DeepMesh, gaps of $21.0$, $28.5$, and $38.8$ percentage points that exceed those on UK Biobank. The two learned methods are the most affected by the change of cohort: CPD-Net falls behind plain CPD for every metric except LV DSC, while on UK Biobank the two remain close, and DeepMesh loses $22.9$ percentage points of Myo DSC and $18.6$ of RV DSC between the datasets, against $5.3$ and $4.2$ for CardioDrift.

\begin{figure}[!t]
    \centering
    \includegraphics[width=\columnwidth]{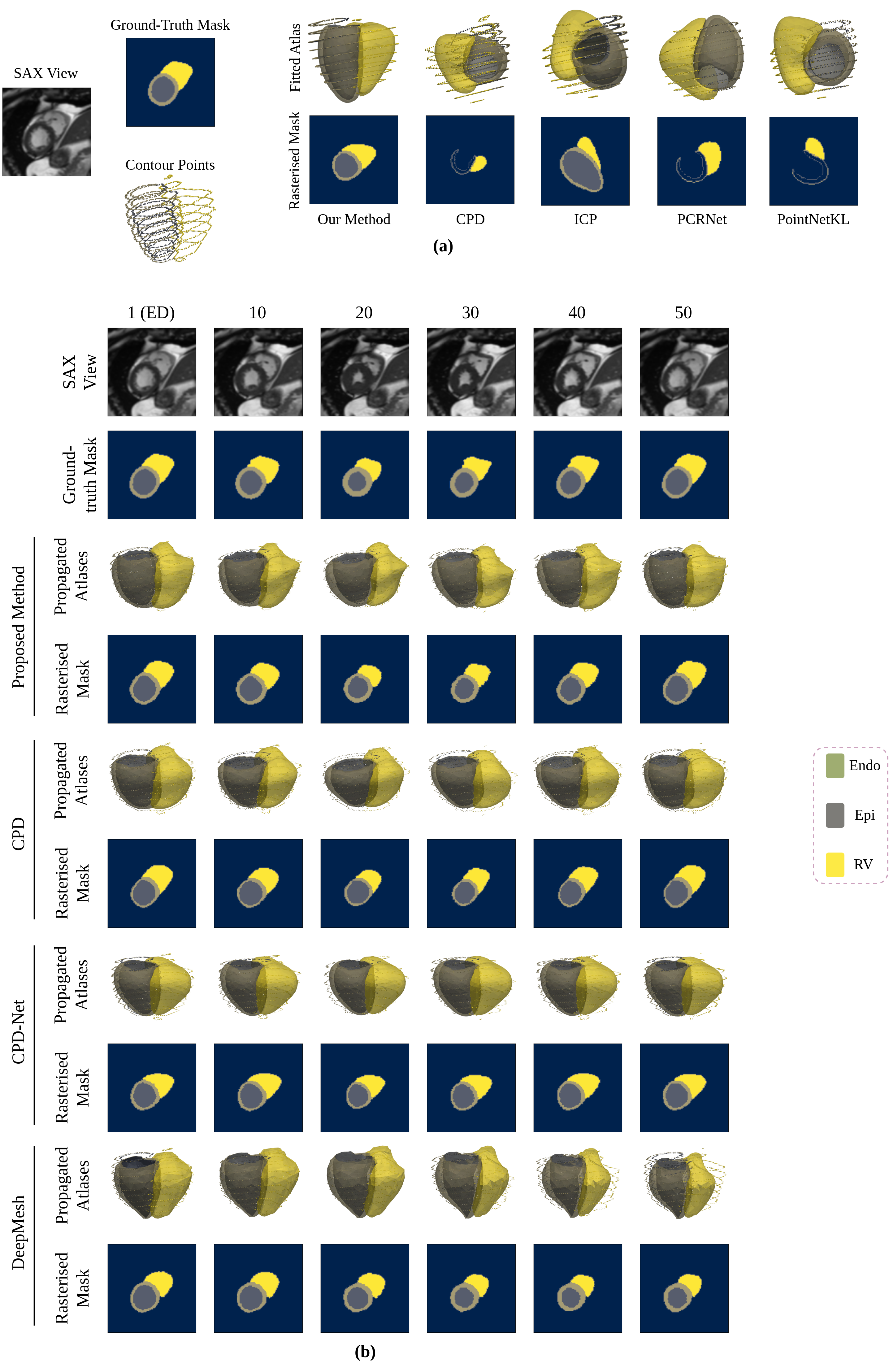}
    \caption{\fontsize{8pt}{10pt}\selectfont Qualitative results. (a) Rigid atlas fitting for CardioDrift and the four baselines, with the mid-ventricular SAX image, ground-truth mask, and extracted contour points on the left; per method, the fitted atlas over the contour points and the rasterised mask. (b) Motion tracking across the cardiac cycle for CardioDrift, CPD, CPD-Net, and DeepMesh; per method, the propagated atlas and its rasterised mask, against the SAX image and ground-truth mask.}
    \label{fig_re}
\end{figure}

Fig.~\ref{fig_re}(b) shows a representative UK Biobank tracking result. The atlases propagated by CardioDrift stay closer to the contour points throughout the cycle and delineate the ED-to-ES transition more faithfully than the three baselines, most visibly in the Myo at ES (frames 20 and 30), indicating that it captures the large inward displacement of the myocardial wall while keeping the mesh smooth and anatomically plausible.

\subsection{Slice-Shift Correction Performance}
\begin{figure}[!b]
    \centering
    \includegraphics[width=0.5\columnwidth]{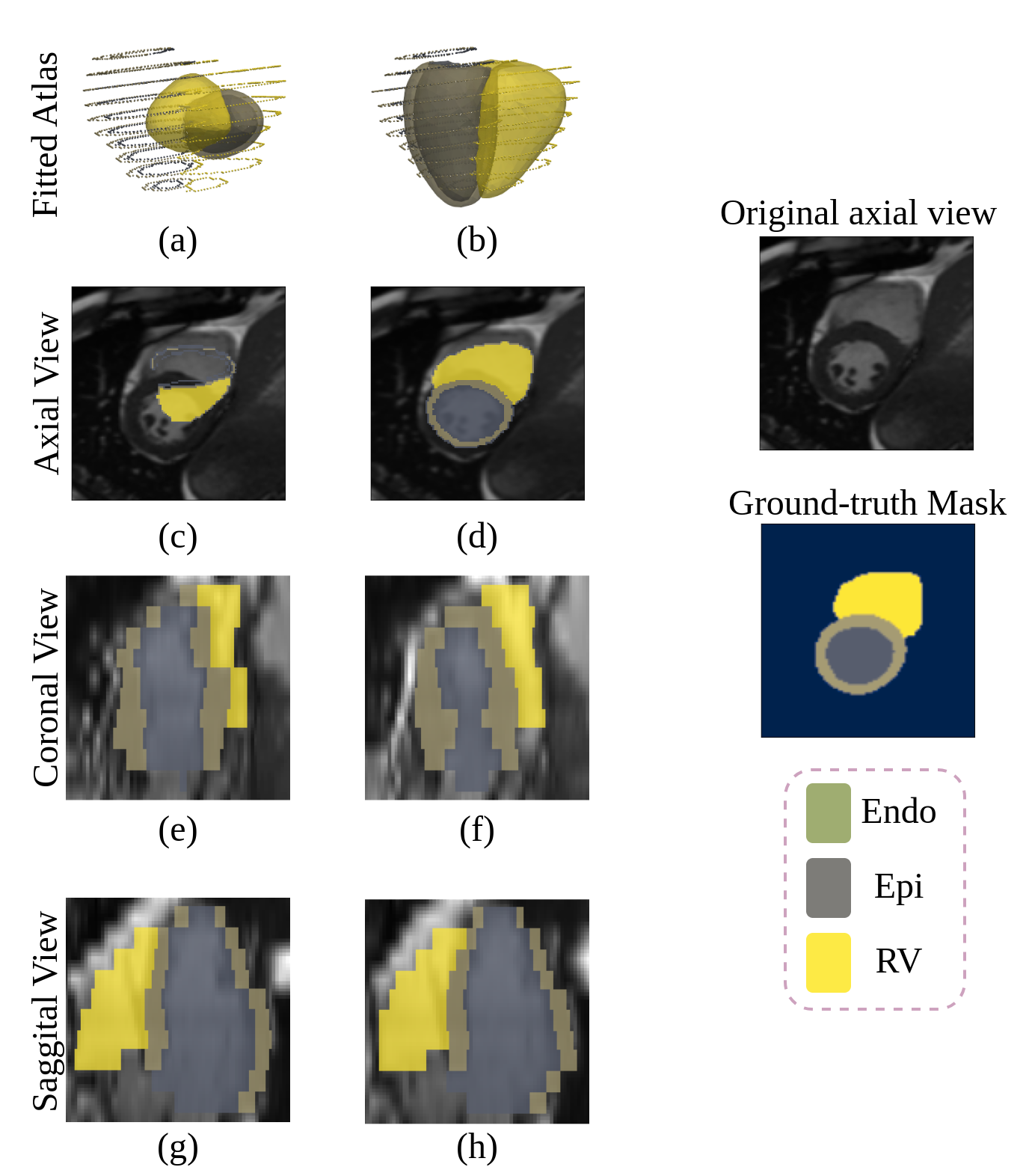}
    \caption{Slice-shift correction. Atlas fitted (a) with and (b) without correction, over the subject's contour points. Axial view of the original middle slice SAX image overlaid with the rasterised mask (c) with and (d) without slice shift correction. Coronal (e,~f) and Sagittal (g,~h) views of the original and corrected SAX images, with ground-truth masks overlaid.}
    \label{fig_SSC_results}
\end{figure}

The two CardioDrift rows of the UK Biobank rigid block in Table~\ref{tab:results} isolate the effect of slice-shift correction. Enabling it improves every LV and Myo metric, most notably Myo DSC ($55.76\pm12.68\%$ to $70.25\pm12.27\%$), while the three RV metrics are marginally better without it. These population-level differences are modest, but the subject-level effect matters: without correction the fitted atlas is visibly offset in a subset of cases (Fig.~\ref{fig_SSC_results}(a)--(d)), an error that would propagate into the non-rigid stage.

The corresponding M\&M rows show a weaker but consistent effect. Correction improves seven of the nine metrics, again with Myo DSC the most responsive ($44.45\pm18.56\%$ to $60.21\pm19.10\%$); the exceptions are RV HD95 and RV MSD, marginally lower without correction, matching the pattern on UK Biobank. The reduced benefit follows from the acquisition protocols: UK Biobank stacks are acquired under a standardised sequential breath-hold protocol in which misalignment is systematic and well described by one in-plane translation per slice, whereas the multi-protocol M\&M stacks leave more residual misalignment outside the model.

Figs.~\ref{fig_SSC_results}(e)--(h) show coronal and sagittal views of the original and corrected SAX images. In the sagittal view the Endo region at the apex, discontinuous in the original data, becomes connected after correction, though a minor disconnection remains at the base; in the coronal view, correction removes inter-slice discontinuities in the Endo and RV. The estimated translations therefore reduce the misalignment artefact in the stack itself.

\subsection{Shape-aware Ablation Study}
Table~\ref{tab:ablation_shape_aware} reports the shape-aware ablation. Enabling the terms for all regions provides the best atlas fitting on both datasets: the best overall HD95 of $17.82\pm3.84$~mm and overall DSC of $80.66\pm9.84\%$ on UK
Biobank, and the best overall HD95 of $22.13\pm6.65$~mm, the best Myo DSC, and the best overall DSC of $73.02\pm18.07\%$ on M\&M. The region-specific configurations confirm that each term benefits its corresponding structure, with the same ordering on both datasets: the Endo-only configuration obtains the lowest Endo and Epi HD95 and the highest LV DSC, and on UK Biobank the highest
Myo DSC as well, whereas the RV-only configuration achieves the best RV HD95 and RV DSC. Disabling a region's term degrades that region most.

On M\&M, the two configurations without the RV term raise RV HD95 from $27.00\pm8.62$~mm to $41.70\pm12.78$ (Endo only) and $48.48\pm15.88$ (Epi only), and reduce RV DSC from $75.01\pm18.54\%$ to $38.85\pm16.37\%$ and $34.60\pm15.68\%$; conversely, retaining only the RV term raises Endo HD95 to $31.88\pm17.30$~mm and lowers LV DSC to $48.92\pm37.16\%$. Because no single-region configuration holds up across all regions, their joint incorporation is the most robust as the overall
columns reflect on both datasets.

\begin{table*}[t!]
\caption{Ablation study of the shape-aware modelling across anatomical regions (rigid stage), evaluated over UK Biobank and M\&M subjects. \cmark~and \xmark~denote whether the shape-aware term is enabled for the corresponding region. All metrics are reported as Mean$\pm$Std. \textbf{Bold}: best per column within each dataset.}
\label{tab:ablation_shape_aware}
\centering
\renewcommand{\arraystretch}{1.3}
\resizebox{\textwidth}{!}{%
\begin{tabular}{c l c c c c c c c c c c c}
\hline\hline
\multirow{2}{*}{Dataset} & \multirow{2}{*}{Config} & \multirow{2}{*}{\textit{Endo}} & \multirow{2}{*}{\textit{Epi}} & \multirow{2}{*}{\textit{RV}} & \multicolumn{4}{c}{HD95 (mm)} & \multicolumn{4}{c}{DSC(\%)} \\
\cmidrule{6-13}
{} & {} & {} & {} & {} & Endo & Epi & RV & All Regions & LV & Myo & RV & All Regions \\
\hline
\multirow{4}{*}{UK Biobank}
        & Full & \cmark & \cmark & \cmark &
            14.18$\pm$4.43 & 16.98$\pm$5.06 & 22.30$\pm$4.78 & \textbf{17.82$\pm$3.84} &
            90.72$\pm$10.56 & 70.25$\pm$12.27 & 81.01$\pm$9.28 & \textbf{80.66$\pm$9.84} \\
        & Endo only & \cmark & \xmark & \xmark &
            \textbf{11.79$\pm$3.30} & \textbf{15.84$\pm$4.56} & 37.38$\pm$7.93 & 21.67$\pm$4.01 &
            \textbf{92.59$\pm$4.72} & \textbf{72.29$\pm$6.43} & 35.03$\pm$10.18 & 66.64$\pm$5.33 \\
        & Epi only & \xmark & \cmark & \xmark &
            15.67$\pm$4.98 & 16.33$\pm$4.71 & 39.37$\pm$9.56 & 23.79$\pm$5.29 &
            87.54$\pm$13.43 & 66.92$\pm$13.33 & 33.51$\pm$10.06 & 62.66$\pm$10.36 \\
        & RV only & \xmark & \xmark & \cmark &
            41.00$\pm$17.61 & 41.24$\pm$16.52 & \textbf{17.19$\pm$3.21} & 33.14$\pm$11.68 &
            55.13$\pm$37.03 & 30.56$\pm$20.94 & \textbf{81.46$\pm$9.44} & 55.72$\pm$21.86 \\
\hline
\multirow{4}{*}{M\&M}
        & Full & \cmark & \cmark & \cmark &
            17.23$\pm$6.98 & 22.15$\pm$7.29 & 27.00$\pm$8.62 & \textbf{22.13$\pm$6.65} &
            83.82$\pm$20.34 & \textbf{60.21$\pm$19.10} & 75.01$\pm$18.54 & \textbf{73.02$\pm$18.07} \\
        & Endo only & \cmark & \xmark & \xmark &
            \textbf{15.85$\pm$6.06} & \textbf{21.31$\pm$6.82} & 41.70$\pm$12.78 & 26.28$\pm$7.42 &
            \textbf{84.92$\pm$19.04} & 58.76$\pm$18.02 & 38.85$\pm$16.37 & 60.84$\pm$14.90 \\
        & Epi only & \xmark & \cmark & \xmark &
            18.94$\pm$5.90 & 23.98$\pm$7.50 & 48.48$\pm$15.88 & 30.47$\pm$8.47 &
            67.48$\pm$29.26 & 47.07$\pm$23.44 & 34.60$\pm$15.68 & 49.71$\pm$20.79 \\
        & RV only & \xmark & \xmark & \cmark &
            31.88$\pm$17.30 & 35.00$\pm$16.32 & \textbf{21.08$\pm$3.97} & 29.32$\pm$11.55 &
            48.92$\pm$37.16 & 29.13$\pm$22.22 & \textbf{79.04$\pm$9.49} & 52.36$\pm$22.12 \\
\hline\hline
\end{tabular}%
}
\end{table*}

\section{Discussion and Conclusion}
This study introduced CardioDrift, a fully automatic, atlas-based framework for representing subject-specific cardiac motion on a 3D atlas mesh with globally corresponding points, requiring only cine CMR SAX segmentation maps. Its core design generalises CPD by jointly registering the Endo, Epi, and RV atlas surfaces to their anatomically corresponding contours under a shared transformation, establishing correspondence within each region rather than across an undifferentiated point set. Moreover, it incorporates slice-shift correction during atlas fitting to the subject’s cardiac space, reducing inter-slice misalignment and improving atlas-fitting accuracy. Compared with state-of-the-art methods, CardioDrift represents subject-specific cardiac motion on an atlas with globally corresponding points, without requiring manual atlas alignment in the subject’s cardiac space. Moreover, it avoids capturing irrelevant motion and the need for image-to-mesh interpolation, reducing attenuation of the derived motion because it operates entirely in the mesh/point-set domain and is based on cardiac contours.

CardioDrift outperformed all baselines on every metric of both datasets. In the rigid stage, the probabilistic formulation expresses correspondence as a soft assignment refined iteratively, suiting the sparse, noisy, and heterogeneous cardiac contour sets, precisely the conditions under which single-pass learning-based methods perform poorly. Embedding slice-shift correction in the same EM procedure reduces the inter-slice misalignment that would otherwise bias the alignment. These translations are estimated only at ED and held fixed thereafter, since the full cine sequence for a slice is acquired within a single breath-hold and the misalignment arises from diaphragm position varying between breath-holds, that is, between slices \cite{petersen2015uk,chandler2008correction}; it is an artefact of acquisition rather than cardiac phase. Unlike DeepMesh \cite{meng2023deepmesh}, the closest prior work, which requires semi-automatic initialisation and interpolates image-space motion fields onto the mesh, CardioDrift automatically fits the atlas to the subject's cardiac space using only the contour points, removing both manual intervention and the associated motion attenuation.

In the non-rigid stage, the largest gain was in Myo DSC, the most difficult region to track owing to its thin structure and large inward displacement at ES. The shape-aware mechanism is decisive: treating the three regions as distinct within a shared deformation field resolves correspondence anatomically rather than geometrically. The comparison with DeepMesh is the most direct of the three baselines, since it likewise propagates a corresponding atlas mesh but differs in how deformation is obtained: its motion fields are estimated from the image domain and interpolated onto the mesh space, and its DSC is $11.0$ percentage points below CardioDrift for the LV and $12.2$ for the RV, indicating that image-to-mesh motion interpolation degrades tracking.

The M\&M evaluation tests these mechanisms outside the standardised UK Biobank acquisition. Likewise UK Biobank, CardioDrift obtains the best rigid and non-rigid registration performance among the baselines, whereas the learned methods lose most, CPD-Net falling below CPD and DeepMesh, which additionally depends on vendor-dependent image intensities, showing the largest drop of any non-rigid method. Because correspondence is re-estimated per subject by EM rather than predicted from learned features, CardioDrift has no training distribution to depart from, and its behaviour is governed by the geometry of the contours presented to it; the regularisation coefficients, selected on UK Biobank and retained unchanged for M\&M, support the same point. The one mechanism whose benefit diminishes is slice-shift correction, since a single in-plane translation per slice describes systematic breath-hold misalignment more completely than the heterogeneous residual misalignment of a multi-protocol cohort.

Several limitations motivate future work. The first is computational cost: \eqref{eq21} must be solved independently at each frame; replacing the iterative EM with a learned model is a natural remedy, though a hybrid coupling a network with a few EM iterations is more promising than a single forward pass, given the sparsity and heterogeneity of cardiac contours. Second, the method depends on the input segmentation, whose errors propagate directly into the contour points; jointly refining the contours from cine CMR image features, or modelling segmentation uncertainty end-to-end, would reduce this sensitivity. The third is the scope of the M\&M evaluation, where reference masks exist only at ED and ES, so temporal consistency over the intermediate frames rests on UK Biobank only. Finally, the formulation includes no term controlling the temporal smoothness of the output mesh sequences. Addressing these would extend the accuracy and anatomical interpretability demonstrated here into a computationally efficient framework for large-scale inter-subject cardiac motion assessment.

\section*{Code Availability}
The source code is publicly available at
\href{https://github.com/rezamovahed93/CardioDrift.git}{GitHub}.

\bibliographystyle{IEEEtran}
\bibliography{ref}

\end{document}